\documentclass[letterpaper, 10 pt, conference]{ieeeconf}  

\IEEEoverridecommandlockouts                              

\usepackage{amsmath} 
\usepackage{amsfonts}
\usepackage{algorithm}
\usepackage{algpseudocode}
\usepackage{graphicx}
\usepackage{xcolor}
\usepackage{color}
\usepackage{url}

\graphicspath{ {./images/} }

\algtext*{EndFor}
\algtext*{EndIf}
\algtext*{EndProcedure}

\renewcommand{\baselinestretch}{0.99}

\title{\LARGE \bf Learned Risk Metric Maps for Kinodynamic Systems}

\author{Ross E.~Allen$^{1}$, Wei Xiao$^{2}$, and Daniela Rus$^{2}$
\thanks{Research was sponsored by the United States Air Force Research Laboratory and the United States Air Force Artificial Intelligence Accelerator and was accomplished under Cooperative Agreement Number FA8750-19-2-1000. The views and conclusions contained in this document are those of the authors and should not be interpreted as representing the official policies, either expressed or implied, of the United States Air Force or the U.S. Government. The U.S. Government is authorized to reproduce and distribute reprints for Government purposes notwithstanding any copyright notation herein.}
\thanks{© 2023 IEEE.  Personal use of this material is permitted.  Permission from IEEE must be obtained for all other uses, in any current or future media, including reprinting/republishing this material for advertising or promotional purposes, creating new collective works, for resale or redistribution to servers or lists, or reuse of any copyrighted component of this work in other works}
\thanks{$^{1}$Massachusetts Institute of Technology, Lincoln Laboratory, Lexington, MA, 02421, USA {\tt\small ross.allen@ll.mit.edu}}%
\thanks{$^{2}$Massachusetts Institute of Technology,    Computer Science and Artificial Intelligence Laboratory,   Cambridge, MA, 02139, USA     {\tt\small weixy@mit.edu; rus@csail.mit.edu}}%
}

\begin{document}

\maketitle
\thispagestyle{empty}
\pagestyle{empty}

\begin{abstract}
    We present Learned Risk Metric Maps (LRMM) for real-time estimation of coherent risk metrics of high-dimensional dynamical systems operating in unstructured, partially observed environments. %
    LRMM models are simple to design and train---requiring only procedural generation of obstacle sets, state and control sampling, and supervised training of a function approximator---which makes them broadly applicable to arbitrary system dynamics and obstacle sets. %
    In a parallel autonomy setting, we demonstrate the model's ability to rapidly infer collision probabilities of a fast-moving car-like robot driving recklessly in an obstructed environment; allowing the LRMM agent to intervene, take control of the vehicle, and avoid collisions. %
    In this time-critical scenario, we show that LRMMs can evaluate risk metrics 20-100x times faster than alternative safety algorithms based on control barrier functions (CBFs) and Hamilton-Jacobi reachability (HJ-reach), leading to 5-15\% fewer obstacle collisions by the LRMM agent than CBFs and HJ-reach. %
    This performance improvement comes in spite of the fact that the LRMM model only has access to local/partial observation of obstacles, whereas the CBF and HJ-reach agents are granted privileged/global information. %
    We also show that our model can be equally well trained on a 12-dimensional quadrotor system operating in an obstructed indoor environment. %
    The LRMM codebase is provided at \url{https://github.com/mit-drl/pyrmm}.
\end{abstract}

\section{INTRODUCTION}
\label{sec:intro}

The estimation of risk is a topic that spans many research domains: from medical treatments \cite{fatemi2021medical}, to economics and portfolio optimization \cite{artzner1999coherent,wang2000class}, to robotics and autonomous systems \cite{majumdar2020should}. %
Within the field of autonomous systems, we often consider risk in terms of probability that a system causes harm to itself or its environment, and the expected cost of such events. 
In this work we seek to estimate the probability of a dynamical system arriving at a state of failure given an initial configuration and control policy. %

\begin{figure}
    \centering
    \includegraphics[width=0.4\textwidth,keepaspectratio]{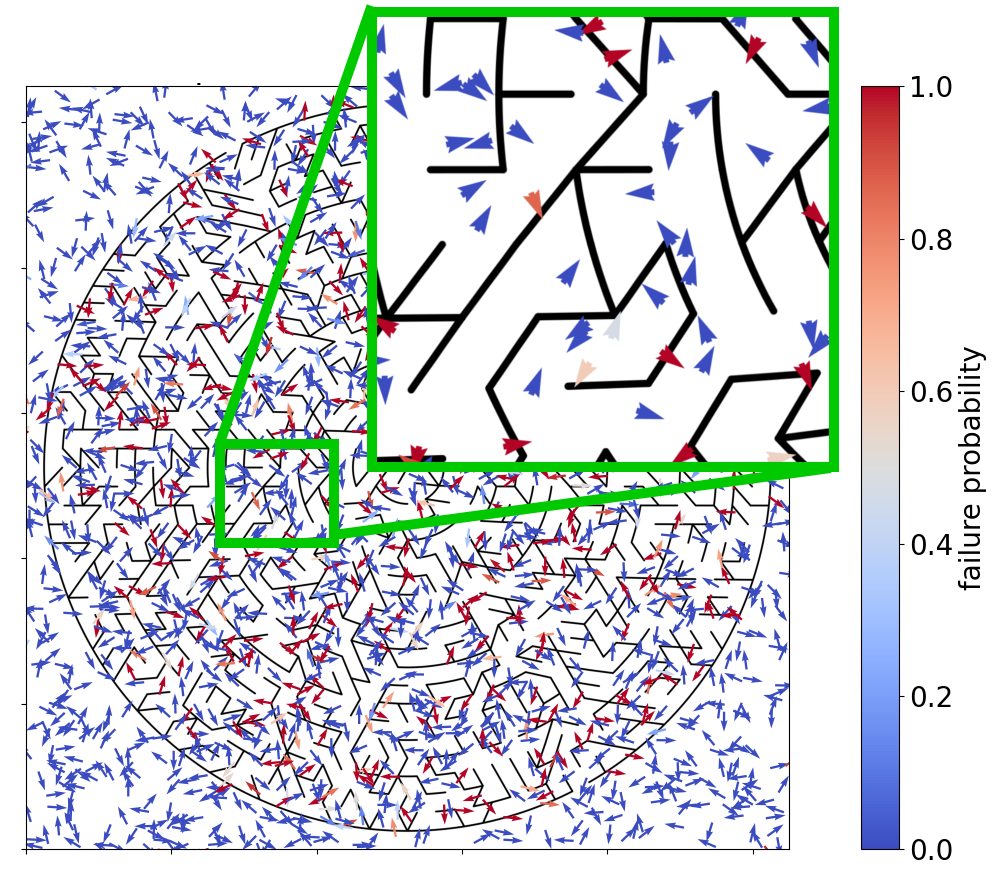}
    \caption{Visualization of inferred risk metrics for 1000 sampled states of a Dubins vehicle within a procedurally generated maze \cite{razimantv2022mazegen}. The minimum turning radius of the vehicle is $\sim$10 pixels. The estimated probability of collision with walls for a random control policy over a finite time horizon is colorized. We see that states with an imminent wall collision are correctly estimated as high risk, whereas unobstructed states are low risk of collision. For clarification, the base of each arrow, not its tip, represent the exact \emph{xy}-coordinate of the sampled state. 
    }
    \label{fig:dubins_maze_risk_metrics}
\end{figure}

Such risk estimation problems have been a major focus within autonomous automobile research where collision avoidance is of utmost concern \cite{lefevre2014survey}. Many of these works emphasise prediction of future states of other vehicles \cite{wang2020fast,pierson2018navigating,hakobyan2021distributionally}. %
These works often adopt domain-specific assumptions that make it difficult to extend techniques beyond the automotive domain---such as strict assumptions on vehicle dynamics\cite{pierson2018navigating}; the assumption that all obstacles or other agents can be represented as points \cite{pierson2018navigating}, balls \cite{hakobyan2021distributionally}, or ellipsoids \cite{wang2020fast}; and assumptions that other agents act in predictable \cite{wang2020fast,hakobyan2021distributionally} and/or self-preserving \cite{pierson2018navigating} fashions.

A closely related problem is that of trajectory collision probability assessment \cite{janson2018monte,schmerling2016evaluating} and risk-aware motion planning \cite{janson2018monte,jasour2019risk,blackmore2010probabilistic}. These works evaluate the probability that a particular trajectory intersects the obstacle space given bounded process and/or measurement noise. 
This is a related---yet \emph{distinct}---problem from the one considered in this paper: i.e. evaluating the probability that a system will arrive at a future failure state given some initial state and control policy. %
Furthermore risk-aware motion planning algorithms almost all rely on an \emph{a priori} map of the obstacle space with, at most, bounded uncertainty on obstacle locations \cite{jasour2019risk}. %
Therefore, even though our work shares concepts and keywords with the field of sampling-based motion planning, it occupies a different problem space than algorithms like rapidly-exploring random tree (RRT)~\cite{lavalle2006planning}.

Hamilton-Jacobi reachability (HJ-reach) analysis \cite{bansal2017hamilton} and control barrier functions (CBFs) \cite{ames2016control}---both of which are discussed more thoroughly in Sec. \ref{sec:related_work} and \ref{sec:experiments}---offer rigorous methods for estimating unsafe sets and providing safety-guaranteed controls for autonomous systems. %
However, the former suffers from limitations on scaling to high-dimensional systems (i.e the ``curse of dimensionality'') and the latter lacks generalizability to arbitrary system dynamics and obstacle sets. %
The scalability limitations of HJ-reachability and generalizability limitations of CBFs motivate the need for alternative methods for assessing safety of dynamical systems and discovering safe control policies. %
Drawing from the field of motion planning, sampling-based methods are a well-known technique for overcoming the curse of dimensionality\cite{lavalle2006planning}. Sampling-based techniques can be applied to systems with arbitrary dynamics and obstacle spaces since these properties are inherent to the sampling process. %
The primary drawback of sampling-based techniques to safety-critical applications is that they may lack the strict safety guarantees that alternative methods provide. %
Lew et al. \cite{lew2022simple,lew2020sampling} provide strong justification for the use of sampling-based methods for safety-critical applications by proving asymptotic convergence to a conservative over-approximation of reachable sets using random set theory. 

In this paper we provide a simple, sampling-based method for approximating risk metrics for arbitrary dynamical systems in partially-observed environments and train a supervised-learning model---referred to as learned risk metric maps (LRMM)---that can rapidly infer failure probabilities in \emph{a priori} unknown environments. We then show that, in spite of HJ-reach and CBF safety guarantees, the LRMM model can provide a greater level of safety in a time-critical parallel autonomy task due to it's rapid estimation of risk. 
    

\section{RELATED WORK}
\label{sec:related_work}

In this section we position our work relative to the closely related techniques of Hamilton-Jacobi reachability (HJ-reach) and control barrier functions (CBF).

The theoretical foundations for this paper can be traced back to the concept of inevitable collision states; i.e. the set of states for which a dynamical system cannot avoid future collisions with obstacles regardless of control input \cite{fraichard2004inevitable,bautin2010inevitable}. 
HJ-reachability represents a modern treatment of inevitable collision states that provides a rigorous theoretical foundation for computing the set of unsafe states that may lead to failure (e.g. collision); often referred to as the ``backward reachable set'' (BRS) \cite{bansal2017hamilton}. %

The BRS of the obstacle space, $\mathcal{C}_{\mathtt{obs}}$, is computed by solving for the value function, $V(t,s_\omega)$, of the Hamilton-Jacobi-Isaacs partial differential equation (PDE) for all states over time horizon $t$ \cite{margellos2011hamilton,fisac2015reach}. %
The value function then allows us to described the BRS of the obstacle space as $\mathcal{V}_{\mathtt{obs}}(t)=\{s : V(t,x) \leq 0 \}$. This is the set---often called the \emph{unsafe set}---for which there exists a control action, $a$ that would lead the system to collision with the obstacle space over time horizon $t$. The HJ value function also provides a means to calculate the optimal control for avoiding obstacles as a gradient of the value function; see Bajcsy et al. for details \cite{bajcsy2019efficient}.

HJ-reachability has been used to develop provably-safe autonomous navigation algorithms for arbitrary dynamical systems within partially-observed environments by treating the unknown portion of the environment as an obstacle \cite{bajcsy2019efficient}.
While the general applicability of HJ-reachability makes it a powerful tool, it suffers from the ``curse of dimensionality'' that make it very difficult to apply the technique in real-time on anything save the simplest, low-dimensional, slow-moving systems \cite{bansal2017hamilton,choi2021robust, bajcsy2019efficient}. %


Control Barrier Functions (CBF) offer an alternative, complementary approach to HJ-reachability that provide provably-safe control of autonomous systems in the presence of obstacles \cite{ames2016control,xiao2021high}. %
In contrast to HJ-reachability, CBFs can provide real-time safe control for high-dimensional control-affine systems. %
CBF-based controllers---or more precisely, controllers based on CBFs \emph{and} control Lyapunov functions (CLFs)---work by mapping state (safety) constraints onto a set of control constraints by taking Lie derivatives of the constraints along the dynamics, and then encoding them as inequality constraints within a quadratic program (QP) that can be solved to determine stabilizing (i.e. target-tracking, using CLFs) and safe (i.e. obstacle-avoiding, using CBFs) control inputs; see Ames et al. \cite{ames2016control} for a detailed discussion. %

General methods do not exist for discovering CBFs for a given control system, requiring that they be hand-designed. %
Furthermore CBFs require an explicit mathematical model of obstacles or keep-out regions to be avoided; something that is often impractical in real-world applications. %
Choi et al. \cite{choi2021robust} propose a method for blending HJ-reachability and CBFs; however this technique still suffers from the poor scalability to high-dimensional systems.

The learned risk metric maps (LRMM) described in the following sections attempt to overcome the drawbacks of HJ-reach and CBF methods by providing a constructive (i.e. not hand-designed) model that rapidly estimates risk metrics for high-dimensional dynamical systems within arbitrary obstacle spaces.

\section{RISK METRIC MAPS}
\label{sec:risk_metric_maps}


Our work considers partially observable Markov decision processes~\cite{kochenderfer2015decision} defined as the tuple  $\left(\mathcal{C}, \mathcal{A}, \mathcal{O}, T, R \right)$. %
$\mathcal{C}$ is the configuration space that defines the possible states of the system, $s \in \mathcal{C}$, as well as the obstacles and/or failure-states of the system, $\mathcal{C}_{\mathtt{obs}} \subset \mathcal{C}$. %
Let the obstacle-free set of the configuration space be defined as $\mathcal{C}_\mathtt{free} = \mathcal{C} \setminus \mathcal{C}_{\mathtt{obs}}$. %
$\mathcal{A}$ is the action space and an action is given as $a  \in \mathcal{A}$. %
$\mathcal{O}(s)$ is the observation function; an observation is given as $o \sim \mathcal{O}(s)$. %
The state transition function, $T \left(s' | s, a \right)$, represents probability of arriving in state $s'$ when taking action $a$ in state $s$. %
The reward is drawn from the reward function $r \sim R(s, a)$. 
A stochastic policy  $\pi (a | o) \in \Pi$ represents the probability, or probability density, of taking action $a$ given observation $o$. %

An observation-action trajectory from timestep $\alpha$ to $\omega$ is defined as $\tau_{\alpha,\omega} = \left(o^{(\alpha)}, a^{(\alpha)}, o^{(\alpha+1)}, a^{(\alpha+1)}, ..., o^{(\omega-1)}, a^{(\omega-1)}, o^{(\omega)}  \right)$. %
Note that we use parenthetical superscripts to represent specific time steps. %
Let $\mathcal{T}$ be the set of all possible trajectories $\tau$ under dynamics $T$ and policy $\pi$. %
Let $J(\tau_{\alpha,\omega}): \mathcal{T} \rightarrow \mathbb{R}$ be the \emph{trajectory cost function} that maps a trajectory to a real-valued number (e.g. fuel usage, time, etc.).


With slight abuse of notation on the variables $\alpha$ and $\omega$, let us define the \emph{cost-limited forward reachable set}~\cite{allen2019real} from state $s_\alpha$ to all states $s_\omega$
\begin{equation}\label{eqn:forward_reachable_set}
    \Omega(s_\alpha,J_{\mathtt{th}}) = \{ s_\omega \in \mathcal{C} | \exists \tau_{\alpha, \omega} \in \mathcal{T}, J(\tau_{\alpha,\omega}) \leq J_{\mathtt{th}}  \} 
\end{equation}

Define the \emph{failure cost function}
\begin{equation}
    Z: \mathcal{C} \rightarrow \mathbb{R}
\end{equation}
that maps each state $s \in \mathcal{C}$ to a real value that corresponds to to the failure-state set $\mathcal{C}_{\mathtt{obs}}$. In this work we define a failure cost function such that
\begin{equation}\label{eqn:failure_cost_func_obstacle}
\begin{split}
& Z(s_\omega \in \mathcal{C}_{\mathtt{obs}}) = 1 \\
& Z(s_\omega \in \mathcal{C}_{\mathtt{free}}) = 0    
\end{split}
\end{equation}
Let $\mathcal{Z}$ be the set of all failure cost functions $Z$. Define the \emph{risk metric} as
\begin{equation}\label{eqn:risk_metric}
    \rho: \mathcal{Z} \rightarrow \mathbb{R}
\end{equation}
In this work we only consider \emph{coherent risk metrics} such as conditional value at risk (CVaR), worst case, or expected cost~\cite{majumdar2020should}. 

Finally---with slight abuse of notation on $\rho$---we can define the \emph{risk metric map} 

\begin{equation}\label{eqn:risk_metric_map}
    \rho \left( s,\pi; Z,\mathcal{C}_{\mathtt{obs}},T \right): \mathcal{C} \times \Pi \rightarrow [0,1]
\end{equation}
which maps the configuration and policy space to the real-value range $[0,1]$ by assigning to each state-policy pair $(s,\pi)$ a risk value parameterized by the failure cost function $Z$, obstacle space $\mathcal{C}_{\mathtt{obs}}$, and vehicle dynamics $T$. By using the binary failure cost function in Eqn. \ref{eqn:failure_cost_func_obstacle} and expected cost as our risk metric, our risk metric map is identical to the \emph{probability} the system arrives at a failure state over an infinite time horizon when starting from state $s$ and following policy $\pi$. %
For brevity of notation we drop the parameter variables and refer to the risk metric map as $\rho(s,\pi)$ or even $\rho(s)$ when the policy is assumed to be uniform random sampling of bounded controls. %
Note that we can relate the risk metric map to \emph{inevitable collision obstacles} (ICO) \cite{fraichard2004inevitable} as

\begin{equation}\label{eqn:region_of_inevitable_collision}
    ICO \left( \mathcal{C}_{\mathtt{obs}} \right) = \{ s \in \mathcal{C} | \forall \pi, \rho(s,\pi) = 1 \}
\end{equation}

\subsection{Approximate Risk Metrics}
\label{sec:apprx_risk_metric_maps}

Many kinodynamic systems of interest---such as cars, aircraft, maritime vessels, etc.---have continuous state and action spaces making enumeration over all states and control trajectories in Eqn. \ref{eqn:forward_reachable_set} impossible. Furthermore, obstacle spaces in Eqn. \ref{eqn:failure_cost_func_obstacle} are often implicitly defined and only sensed with partial observability. This means that the explicit derivation of risk metrics at each state of a system is impractical, if not impossible. %

We can, however, formulate a finite-horizon approximation of Eqn. \ref{eqn:risk_metric_map} at a given state. %
Algorithm \ref{alg:est_risk_metric} gives a recursive, sampling-based estimate of the risk metric at state $s$, which is arrived at by trajectory $\tau$, and subsequently following policy $\pi$ over a time horizon of $t*m$.

\begin{algorithm}[t]
\caption{Approximate Risk Metric} \label{alg:est_risk_metric}
\begin{algorithmic}
\Procedure{ApprxRiskMetric}{$s,\tau,\pi,t,n,m$}
    \State $z\gets\textsc{CheckFailure}(s,\tau)$
    \If {$z=1$ \textbf{or} $m=0$}
        \State \textbf{return} $z$ 
    \EndIf
    \State $P \gets \emptyset$    
    \State $V,E \gets \textsc{SampleForwardReachableSet}(s,\pi,t,n)$
    \For {$s_{\omega}, \tau_{\omega}$ in $\textsc{zip}(V,E)$}
        \State $P \gets P \cup \{\textsc{ApprxRiskMetric}(s_\omega, \tau_\omega, \pi, t, n, m-1) \}$ 
    \EndFor
    \State \textbf{return} $\textsc{CoherentRiskMetric}(P)$ 
\EndProcedure
\end{algorithmic}
\end{algorithm}

The algorithm works by checking if a state is in collision with obstacles (i.e. failure cost function from Eqn. \ref{eqn:failure_cost_func_obstacle}) and then recursively sampling the stochastic policy $\pi$ which samples the forward reachable set $\Omega(s,t)$ of state $s$ with a cost function of time $t$ (Eqn. \ref{eqn:forward_reachable_set}). This recursive sampling generates a tree rooted at $s$ with a branching factor of $n$ and depth of $m$. %
The recursive base of the algorithm is to simply return the value of the failure cost function at the leaf node, which is either 0 or 1. %
Otherwise the algorithm returns a coherent risk metric---such as expected cost---over all immediately adjacent sampled states in the tree.



\subsection{Learned Risk Metric Maps}
\label{sec:learned_risk_metric_maps}

While Alg. \ref{alg:est_risk_metric} provides an approximation to risk metrics in Eqn. \ref{eqn:risk_metric_map}, this only provides an approximation at a single state and not an approximation over the entire state space. Furthermore, we need privileged access to obstacle information in order to perform collision checking during the sampling process. This means that Alg. \ref{alg:est_risk_metric} is of limited use in real-time systems where obstacles may only be partially sensed (e.g. via cameras or Lidar).

We seek a method for approximating risk metrics over finite time horizons for kinodynamic systems operating in unstructured, partially observed obstacle spaces. To achieve this we use Alg. \ref{alg:est_risk_metric} as a data generator to train a function approximator (e.g. neural network) that takes as input the local observation, $o \sim \mathcal{O}(s)$, and outputs the \emph{inferred} risk metric, $\hat{\rho}(s)$. Training data is generated under policy $\pi_g$ (e.g. uniform random from bounded control space \cite{lew2022simple,lew2020sampling}). Therefore we define the \emph{learned risk metric map} (LRMM) function approximator as

\begin{equation}\label{eqn:lrmm}
    \hat{\rho}(o; \pi_g) : \mathcal{O} \rightarrow [0,1]
\end{equation}

Algorithm \ref{alg:train_rist_metric_maps} gives a rough overview of the relatively simple procedure for training the LRMM model in Eqn. \ref{eqn:lrmm}. This consists of generating obstacle sets; sampling states from these obstructed configuration spaces; pulling observations from these states and computing risk metrics at each state; and finally, training a supervised model on these observations and risk metrics. 

\begin{algorithm}[t]
\caption{Training Learned Risk Metric Maps} \label{alg:train_rist_metric_maps}
\begin{algorithmic}
    \State $\mathbf{C}_M  \gets \textsc{GenerateObstacleSets}(M)$
    \State $\mathbf{S}_N \gets \textsc{SampleStates}(\mathbf{C}_M,N)$ 
    \State $\mathbf{O}_N \gets \textsc{ObserveStates}(\mathbf{S}_N)$
    \State $\mathbf{P}_N \gets \textsc{ApprxRiskMetrics}(\mathbf{S}_N,\tau,\pi,t,n,m)$
    \State $\hat{\rho} \gets \textsc{TrainModel} (\text{inputs=}\mathbf{O}_N, \text{targets=}\mathbf{P}_N)$
\end{algorithmic}
\end{algorithm}

Similar to sampling-based motion planners (e.g. RRT~\cite{lavalle2006planning}) and online planning methods (e.g. monte carlo tree search~\cite{kochenderfer2015decision}) approximate risk metrics use sampling techniques to explore the configuration space using tree structures. However, LRMM differs in that it does not directly attempt to determine policies or motion plans for navigating the configuration space. Therefore, it is not necessary to form a connected graph with all sampled states and instead allows for many disjoint trees sampled throughout the configuration space making it highly parallelizable.

\section{EXPERIMENTS}
\label{sec:experiments}

In this section we provide simulation experiments demonstrating the utility of learned risk metric maps. %
We provide a case study on parallel autonomy, comparing effectiveness of LRMMs with control barrier functions (CBFs) and Hamilton-Jacobi reachability (HJ-reach) in a 4-dimensional car-like robot. %
We then provide training results for high-dimensional systems in unstructured obstacles spaces such as a quadrotor in an obstructed room.

For all experiments the LRMM models are trained with Alg. \ref{alg:train_rist_metric_maps} using a branching factor $n=32$, a tree depth $m=2$, time horizon of $t=2$ seconds per tree depth, and expected value of our failure cost function (Eqns. \ref{eqn:failure_cost_func_obstacle}) as our coherent risk metric. 
Similarly, a held-out test set of data is generated. Models consist of a 64-neuron, single layer, feed-forward network with a sigmoid activation function. Another sigmoid function is applied at the output layer to bound outputs to the range $[0,1]$ so that outputs represent probabilities. %
The training process uses a mean squared error (MSE) loss function for the regression problem.

For software implementation, state and control spaces are defined and sampled using Open Motion Planning Library (OMPL) \cite{sucan2012open}. %
Network models and training are implemented with PyTorch \cite{paszke2019pytorch}. 
In Sec. \ref{sec:parallel_autonomy_case_study} CBF quadratic programs are solved using the CVXOPT library \cite{andersen2013cvxopt}. The HJ-reachability PDE is defined and solved with the OptimizedDP library \cite{bui2022optimizeddp}.
In Sec. \ref{sec:high_dim_and_unstructured_exp} PyBullet \cite{coumans2016pybullet} is used for quadrotor system dynamics, 3D collision checking, and rendering. %
All data generation, model training, and evaluation experiment configurations are managed using hydra-zen \cite{soklaski2022tools}. %
LRMM training and experiment software is provided at \url{https://github.com/mit-drl/pyrmm}.


\subsection{Parallel Autonomy Case Study}
\label{sec:parallel_autonomy_case_study}

\begin{figure}
    \centering
    \includegraphics[width=0.28\textwidth]{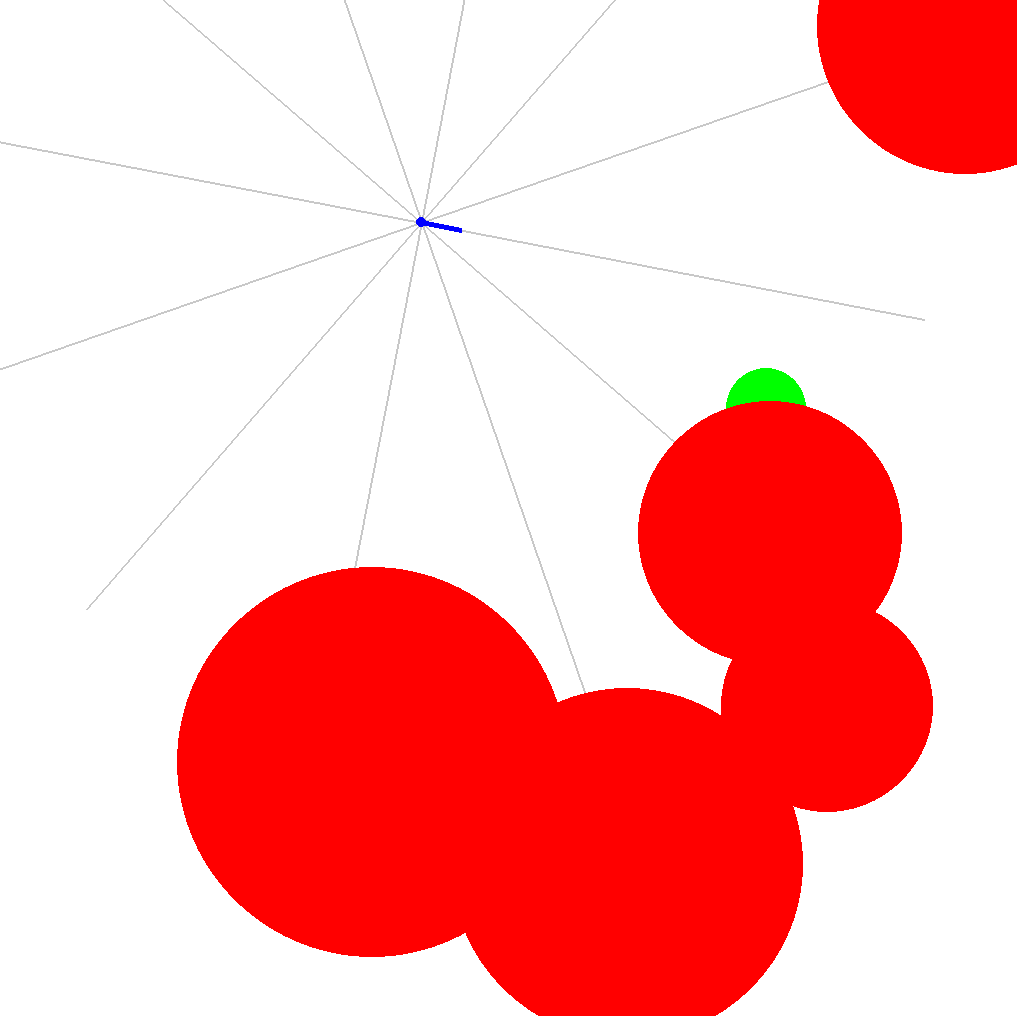}
    \caption{Dubins4D parallel autonomy simulation environment. Obstacles are randomly placed red circles, the goal is the partially obscured green circle, the agent and its velocity vector is indicated in blue, and the lidar ray casts are light gray. A ``reckless'' control system navigates the vehicle to the goal without regard to obstacles. We evaluate LRMM, CBF, and HJ-reach methods as ``guardian agents'' who can take temporarily take control of the vehicle to avoid obstacles. 
    }
    \label{fig:dubins4d_sim_env}
\end{figure}

\textbf{Problem setup.} We consider a Dubins-like kinodynamic system---referred to as Dubins4D and illustrated in Fig. \ref{fig:dubins4d_sim_env}---which is commonly studied in related works \cite{bajcsy2019efficient,xiao2021high} with dynamics given as
\begin{equation}\label{eqn:dubins4d_dynamics}
\begin{split}
    \dot{x} = v\cos\theta + d_x, \ 
    \dot{y} = v\sin\theta + d_y, \ 
    \dot{\theta} = u_1, \ 
    \dot{v} = u_2
\end{split}
\end{equation}
where state, $s:=(x,y,\theta,v)$, represents the location in Cartesian frame, heading, and linear speed of the vehicle, respectively; and controls, $\mathbf{u}:=(u_1, u_2)$, are the turn-rate and linear acceleration, respectively. %
The system is subject to state constraints $v_{\text{min}} \leq v \leq v_{\text{max}}$; control constraints $u_{1,\text{min}} \leq u_1 \leq u_{1,\text{max}}$, $u_{2,\text{min}} \leq u_2 \leq u_{2,\text{max}}$; and bounded disturbance $\|d_x\|\leq d_r, \|d_y\| \leq d_r$.

\begin{table*}[t]
\centering
\caption{Comparative study of learned risk metric maps, control barrier functions, and Hamilton-Jacobi reachability as parallel autonomy agents in the Dubins4D driving environment}
\label{tab:dubins4d_compare_study}
\begin{tabular}{llcrrrrr}
\textbf{Agent} &
  \textbf{\begin{tabular}[c]{@{}l@{}}Observation\\ Type\end{tabular}} &
  \textbf{\begin{tabular}[c]{@{}c@{}}Num.\\ Obstacles\end{tabular}} &
  \textbf{\begin{tabular}[c]{@{}r@{}}Policy \\ Compute\\ Time {[}ms{]}\end{tabular}} &
  \textbf{\begin{tabular}[c]{@{}r@{}}Success \\ Rate \\ {[}\% eps{]}\end{tabular}} &
  \textbf{\begin{tabular}[c]{@{}r@{}}Collision\\ Rate\\ {[}\% eps{]}\end{tabular}} &
  \textbf{\begin{tabular}[c]{@{}r@{}}Timeout\\ Rate\\ {[}\% eps{]}\end{tabular}} &
  \textbf{\begin{tabular}[c]{@{}r@{}}Intervention\\ Rate\\ {[}\% steps{]}\end{tabular}}  \\ \hline \vspace{-3pt} \\ 
LRMM        & local/partial & 5 & \textbf{2.4$\pm$5.0} & $45.3\pm7.3$ & \textbf{29.7$\pm$5.6} & $25.0\pm4.8$ &  $30.7\pm32.5$  \\
CBF         & global/privileged & 5 & $52.9\pm41.2$ & $46.2\pm5.9$ & $35.1\pm5.2$ & $18.8\pm3.8 $ &  $32.8\pm39.4$  \vspace{4pt}\\
random      & none & 5 & $1.4\pm2.7$ & $27.1\pm6.9$ & $68.4\pm5.9$ & $4.5\pm2.9$ & $\sim50$ \\
inactive    & none & 5 & $1.0\pm2.3$ & $44.4\pm7.4$ & $55.4\pm7.3$ & $0.1\pm0.5$ & $0$ \\
brakes-only & none & 5 & $0.3\pm0.7$ & $0.6\pm0.9$ & $9.0\pm3.8$ & $90.4\pm3.7$ & $100$ \vspace{4pt}\\
HJ-reach    & global/privileged & 1 & $344.0\pm850.0$ & $54.0\pm5.2$ & $46.0\pm5.3$ & $0.0\pm0.3$ & $<1.0$ 
\end{tabular}
\end{table*}

We consider the case where the Dubins4D system in Eqn. \ref{eqn:dubins4d_dynamics} is operated in the presences of obstacles and steered by a ``reckless driver'', i.e. a controller that guides the system to a goal region but makes no effort to avoid obstacles\footnote{For our experiments the ``reckless driver'' is a control Lyapunov function (CLF) taken from \cite{xiao2021high}; however, any other controller could be used so long as it generates a non-zero frequency of obstacle collisions.}. %
The objective is to provide a parallel autonomous system---referred to as the ``guardian agent''---that judiciously takes control from the driver to prevent collisions with obstacles \cite{amini2019variational}. %
It is desired that the guardian agent be minimally-interfering; it should only take control when collisions are imminent and is not responsible for optimizing navigation to the goal. %

The environment executes in a \emph{non-blocking} fashion; that is to say that simulation time continues to progress even while agents are computing their next action. This characteristic of the environment is designed to demonstrate the importance of computation time in safety-critical robot applications. %
The observation spaces is a 17-tuple with simulation time, relative position to goal, absolute heading and speed, and 12 ray-casts equally distributed around the vehicle that sense obstacle locations.

In this experiment we compare three guardian agents: \emph{LRMM}, \emph{CBF}, and \emph{HJ-reach}.



\textbf{LRMM:} The \emph{LRMM agent} is trained with 35,072 samples drawn from 274 randomly generated obstacle configurations that are completely distinct from those configurations used during the evaluations in Table \ref{tab:dubins4d_compare_study}; i.e. LRMM was \emph{not} allowed to train on the evaluation set.
During runtime evaluation, the LRMM agent works by inferring risk from observation of the system's current state. If the inferred risk is greater than a user-defined threshold ($\hat{\rho}(s)>0.85$ in our experiments), then the LRMM agent takes active control of the vehicle and applies maximal braking and maximal turning control until the risk estimate drops below the threshold, at which time control is returned to the driver agent\footnote{Note that this collision avoidance policy is a heuristic and not derived from the LRMM output. Thus, in this case study, LRMM helps to answer the question of \emph{when} to take control, but not \emph{what} control to apply; i.e. the LRMM is \emph{not prescriptive} of control. See Sec.~\ref{sec:conclusion} for discussion of future work on this topic.}.

\textbf{CBF:} The \emph{CBF agent} is based on high-order CBFs described in \cite{xiao2021high}. %
CBF-based controllers are most often considered in the context of fully-autonomous systems, but we can easily adapt them to parallel autonomy systems. %
At each new state that the system encounters, the CBF agent formulates and solves the quadratic program (QP) described in Sec. \ref{sec:related_work}. If the CBF constraints are active in the QP solution---i.e. at least one CBF inequality constraint reaches an equality---then the CBF agent overrides the driver and applies the controls resulting from the QP solution. %
Note that CBFs require an explicit, analytical description of the obstacle space in order to formulate the CBF-based QPs.  
For our experiments we work around this by providing the CBF agent with privileged, global knowledge of the obstacle space that is not given to the LRMM agent. 

\textbf{HJ reachability:} The \emph{HJ-reach agent} is based upon Hamilton-Jacobi reachability \cite{bajcsy2019efficient,bansal2017hamilton}.

As the Dubins4D driver moves the vehicle through the state space, the HJ-reach agent assess the value function at each state and determines if it is within the unsafe set $\mathcal{V}_{\mathtt{obs}}(t)$. If the system arrives in the unsafe set, then the HJ-reach agent takes control and applies the optimal control to steer the system away from the obstacle set. %

Several modifications to the Dubins4D environment were necessary for the HJ-reach agent to be applied. First, like the CBF agent, the HJ-reach agent was granted privileged information about the obstacle space in order to formulate the HJ PDE\footnote{Unlike CBFs, HJ-reachability does not explicitly require analytical descriptions of the obstacle space and can---in principle---work with an implicit description, e.g. a discretized mesh of the obstacle space. However, our implementation of HJ-reachability was constrained by the Python libraries available for solving the HJ PDE \cite{bui2022optimizeddp} which required explicit obstacle descriptions (e.g. circular regions of known radius and position)}. %
Furthermore, we granted the HJ-reach agent a precomputation phase where the agent is allowed to \emph{``peek''} at the complete obstacle configuration prior to environment initiation and then given time to solve for the HJ value function over a discretized mesh of the state space. %
The computation of the HJ value function can take up to 2 minutes on the same computer hardware that solves CBF QPs in 50 milliseconds. This is so slow that---without this precompute phase---the entire simulation time window would have elapsed before the HJ-reach agent even had the chance to interact and control the system. 
Finally, we could only impose a single obstacle for the HJ-reach agent due to limitations of the software library used to implement the HJ solver. 

\textbf{Other Baselines:} We also compare against a set of baseline agents---\emph{random}, \emph{inactive}, \emph{brakes-only}---that have simplistic policies that, respectively, intervene with random controls at random times; never intervene on the drivers actions; and always intervene and apply maximum braking at all times. These baselines help bound the possible performance metrics within the Dubins4D environment.  

\begin{table*}[]
\centering
\caption{Risk metric training and testing results on unstructured environments and high-dimensional systems}
\label{tab:dubins_and_quad_training_results}
\begin{tabular}{llrrrrrrr}
\textbf{Dynamics} &
  \textbf{\begin{tabular}[c]{@{}l@{}}Obstacle\\ Space\end{tabular}} &
  \textbf{\begin{tabular}[c]{@{}r@{}}State\\ Dims\end{tabular}} &
  \textbf{\begin{tabular}[c]{@{}r@{}}Control\\ Dims\end{tabular}} &
  \textbf{\begin{tabular}[c]{@{}r@{}}Observe\\ Dims\end{tabular}} &
  \textbf{\begin{tabular}[c]{@{}r@{}}Train\\ Samples\end{tabular}} &
  \textbf{\begin{tabular}[c]{@{}r@{}}Final Train\\ MSE/MAE\end{tabular}} &
  \textbf{\begin{tabular}[c]{@{}r@{}}Test\\ Samples\end{tabular}} &
  \textbf{\begin{tabular}[c]{@{}r@{}}Test\\ MSE/MAE\end{tabular}} \\ \hline \vspace{-3pt} \\
Dubins-3D &
  Mazes &
  3 &
  1 &
  8 &
  65536 &
  \begin{tabular}[c]{@{}r@{}}3.70e-3\\ 1.83e-2\end{tabular} &
  4096 &
  \begin{tabular}[c]{@{}r@{}}3.38e-2\\ 5.51e-2\end{tabular}
  \vspace{4pt} \\
Dubins-4D &
  Circles &
  4 &
  2 &
  17 &
  35072 &
  \begin{tabular}[c]{@{}r@{}}2.38e-3\\ 3.71e-2\end{tabular} &
  5260 &
  \begin{tabular}[c]{@{}r@{}}2.98e-3\\ 3.68e-2\end{tabular}
  \vspace{4pt} \\
Quadrotor &
  \begin{tabular}[c]{@{}l@{}}Walls \& \\ Polyhedra\end{tabular} &
  12 &
  4 &
  16 &
  65536 &
  \begin{tabular}[c]{@{}r@{}}1.69e-4\\ 7.88e-3\end{tabular} &
  8192 &
  \begin{tabular}[c]{@{}r@{}}4.97e-4\\ 1.13e-2\end{tabular}
\end{tabular}
\end{table*}

\textbf{Comparative Results:} Table \ref{tab:dubins4d_compare_study} provides the experimental results from our Dubins4D parallel autonomy case study. A set of 2048 environment configurations (i.e. placement of obstacles, goal, and initial vehicle state) are randomly generated and each agent is evaluated in each of these environment configurations---except the HJ-reach agent which had it's own set of 2048 environments due to its limitation to single obstacles. %
The agents are evaluated based on their policy computation time, success rate (i.e. proportion of trials that end at the goal state), collision rate (trials that end in collision with obstacles), timeout rate (trials that end in neither goal or collision), and intervention rate (i.e. the percent of time steps for which the guardian agent takes control from the driver). 
Ideally, a guardian agent would have a perfect success rate; a zero collision and timeout rate; and a minimal intervention rate (only takes control at the critical moments to avoid collision and then returns control to the driver); however, such a ``perfect'' guardian agent is not actually possible in this contrived scenario---see discussion below. %

The key insight from Table \ref{tab:dubins4d_compare_study} is that the LRMM agent performs at least as well---if not better than---CBFs and HJ-reach methods in this real-time, safety-critical, parallel autonomy task. This is in spite of the fact that the LRMM agent only has access to a local, partial observation of the environment whereas the CBF and HJ-reach agents are granted privileged global knowledge of the obstacle space. %
The success of the LRMM agent is due in large part to its rapid computation time. %
Needing only to make a feed-forward pass through a shallow neural network in order to estimate risk metrics, the LRMM agent can run approximately 20x faster than the CBF agent---which needs to solve quadratic programs at each step---and $>100$x faster than the HJ-reach agent---which needs to interpolate a value function and its derivative on a 4-dimensional mesh of the state space. %
Even though CBFs and HJ reachability can provide theoretical guarantees about control robustness and optimality, these experiments highlight the practical challenges of their implementations in real-time safety-critical applications.

The LRMM and CBF agents both produce roughly the same success rate of $45\%$. 
LRMM slightly outperforms CBFs with a $30\%$ collision rate. 
The Dubins4D environment is randomly initialized such that collisions are unavoidable in some episodes; i.e. the initial state of the vehicle is already within the region of inevitable collision (Eqn. \ref{eqn:region_of_inevitable_collision}) \cite{fraichard2004inevitable}. %
From the inactive-agent we see that---without intervention from a guardian---the ``natural'' proportion of episodes ending in collision is roughly $55\%$. From the brakes-only agent---which always intervenes and immediately brings the vehicle to a stop---we see that an approximate lower bound on collision rate is $9\%$. 
LRMM produces a higher rate of episode timeouts---with neither goal nor obstacle being reached---which tend to be caused by the guardian agent bringing the vehicle to a complete stop until the end of the episode. 
Both LRMM and CBFs have roughly a $30\%$ intervention rate. 
LRMM and CBF agents significantly out-perform the baseline random agent across all metrics.

Direct comparison with HJ-reach in terms of success, collision, and timeout rate is made difficult by the fact that it was not exposed to the same number of obstacles as other agents. 
HJ-reach results are included to highlight the long computation time relative to other methods. This is also what leads to it's low intervention rate. By the time that the HJ-reach agent has identified that the system has entered an unsafe set and computed the optimal evasion control, the vehicle has already collided with the obstacle. 

\subsection{High-Dimensional Systems \& Unstructured Environments}
\label{sec:high_dim_and_unstructured_exp}

\begin{figure}[thpb]
    \centering
    \includegraphics[width=0.3\textwidth]{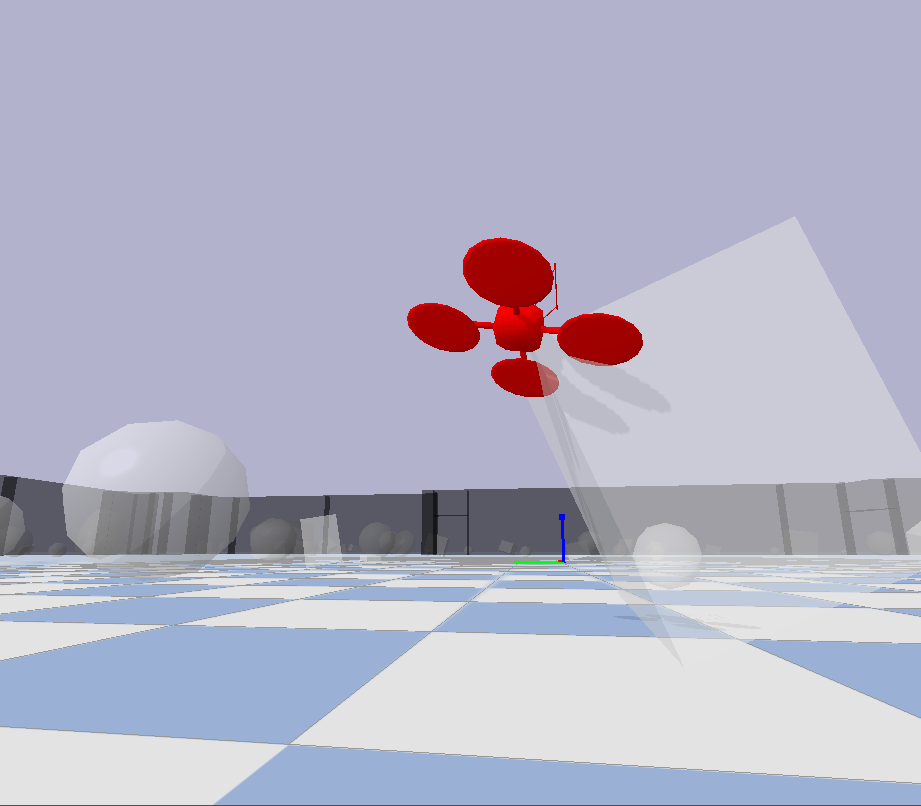}
    \caption{Visualization of quadrotor in an highly unsafe state (i.e. upside down and near an obstacle). Color of the quadrotor is scaled from green-to-red based on the inferred risk metric which, here, accurately estimates the high risk of the current state. Obstacles and walls made transparent for easier visualization. Rendered in PyBullet \cite{coumans2016pybullet}.} 
    \label{fig:quadrotor_risk_viz}
\end{figure}

As demonstrated in Sec. \ref{sec:parallel_autonomy_case_study}, learned risk metric maps can infer a dynamical system's current risk of failure much more rapidly than existing techniques like CBFs and HJ-reachability. %
These alternative techniques become even less usable in higher-dimensional systems---where solutions of PDEs over the state space become prohibitively expensive---and unstructured environments---where obstacles cannot be explicitly/analytically described. %
In this section we show that learned risk metric maps can effectively estimate risk values in such challenging environments. In addition to the Dubins4D environment, we train risk models for a Dubins vehicle in highly-obstructed procedurally-generated mazes (see Fig. \ref{fig:dubins_maze_risk_metrics}) and a quadrotor within procedurally-generated rooms with random polyhedron obstacles (see Fig. \ref{fig:quadrotor_risk_viz}).

Table \ref{tab:dubins_and_quad_training_results} gives the risk metric map training and testing results for these three systems. %
In addition to the MSE training and testing loss we also report mean absolute error (MAE) at the end of training and on the test set. This gives more intuition on how accurately the models estimate risk. We see that all models average a risk estimation error of $<10\%$ of the true risk on the test set.

\section{Conclusion}
\label{sec:conclusion}

In this paper, we present learned risk metric maps (LRMMs) for real-time estimation of coherent risk metrics of high-dimensional dynamical systems operating in unstructured, partially observed environments. We compared the proposed LRMM with other state of the art methods: CBF and HJ-reachability, with results showing advantages of the proposed LRMM's computation efficiency and general applicability. 
As noted in Sec.~\ref{sec:parallel_autonomy_case_study}, a shortcoming of LRMMs is that they estimate failure probabilities, but they do not prescribe the control necessary to avoid failure. %
We anticipate that future work will remedy this---perhaps through some fusion with control barrier functions---and seek to provide formal guarantees on LRMM accuracy~\cite{lew2022simple}. 

\bibliographystyle{IEEEtran}
\bibliography{IEEEabrv,root}

\end{document}